\documentclass[10pt,journal]{IEEEtran}
\usepackage{amsmath,amssymb}
\usepackage{booktabs,graphicx}
\graphicspath{{figures/}}
\usepackage{xcolor}
\usepackage[ruled,vlined]{algorithm2e}

\begin{document}
\bstctlcite{BSTcontrol}

\title{Outlier-Robust Variational Bayesian Smoothing with Decoupled Hierarchical Likelihoods}

\author{Arslan Majal and Aamir Hussain Chughtai

\thanks{A. Majal is with Zunitech, LLC, Houston, USA.
A. H. Chughtai is with the Institute of Data Science, University of
Engineering and Technology, Lahore, Pakistan.}

}

\maketitle

\begin{abstract}
We present a decoupled adaptive selective outlier
rejecting (D-ASOR) smoother for nonlinear state-space models
with independent measurement channels and unknown nominal
noise variances. By decoupling the nominal noise variance
from the outlier likelihood, the hierarchical model enables
robust channel-wise noise learning.
Variational expectation--maximization combines nonlinear
Gaussian smoothing with closed-form updates of outlier
probabilities and noise parameters.
Furthermore, we show that the direct influence of extreme
outliers on nominal-variance learning vanishes under fixed
variational quantities and model parameters.
The proposed method admits linear per-iteration
complexity in the measurement dimension.
Experiments demonstrate improved estimation accuracy
with competitive runtimes relative to existing
robust adaptive methods.
\end{abstract}


\section{Introduction}
\label{sec:introduction}

State estimation is a core task in
several applications spanning
engineering, finance, and scientific modeling
\cite{sarkka2023bayesian}. Given the complete observation sequence, smoothing exploits all measurements to estimate
the latent state trajectory \cite{sarkka2010gaussian}. Conventional smoothers rely on accurate stochastic
dynamical models, but measurement noise uncertainty
and outlier contamination can compromise state estimation.
Nominal noise variances may be uncertain due to imperfect
sensor calibration or limited noise characterization,
while outliers may arise from sensor faults, interference,
or multipath propagation \cite{wang2018robust}.

Addressing these issues presents several
challenges. Naively rejecting the entire measurement vector may discard reliable observations when only a subset is contaminated \cite{chughtai2022,li2021robust}. However, modeling component-wise contamination through
latent indicators introduces a combinatorial problem
requiring specialized inference \cite{majal2024}.
Furthermore, distinguishing outlier characteristics from
nominal noise variability complicates their joint
estimation, particularly in heterogeneous sensing systems
with unknown channel-specific noise variances
\cite{zhu2022}.

To mitigate measurement contamination, robust state
estimation methods employ heavy-tailed distributions,
robust residual losses, or threshold-based outlier
rejection \cite{chen2020smoother,aravkin2014,huang2016smoother,bai2021}. While conventional approaches rely on prespecified
residual loss functions, adaptive methods learn these during estimation \cite{chughtai2022}. Variational Bayesian (VB) theory provides computationally attractive tools to devise adaptive robust estimators \cite{blei2017}.  Leveraging VB, selective outlier rejection methods \cite{chughtai2022} and switching observation models \cite{nakabayashi2020}
address contaminated measurements. Adaptive selective outlier rejecting (ASOR) estimators
further learn outlier statistics during inference
\cite{chughtai2024heuristics,majal2025emorf}. However these methods assume known nominal measurement noise statistics.

To relax the known-noise assumption, methods for joint
state and noise covariance estimation have also been
developed \cite{sarkka2009recursive}. To address noise uncertainty and outliers, \cite{li2016vbhaukf} proposed an estimator (VB-HAUKF) that employs Huber's M-estimation theory. Similarly, \cite{zhu2022} developed an estimator (VBAKF-GGM) based on Gaussian--Gamma mixtures that models outliers through covariance scaling while jointly estimating states and nominal noise covariances. 
While these approaches address noise uncertainty and
outliers through robustified updates or covariance scaling,
we investigate an alternative formulation that explicitly
decouples the outlier likelihood from the nominal
noise variance. 

We propose a decoupled
adaptive selective outlier rejecting (D-ASOR) smoother
for nonlinear state-space models with independent
measurement channels and unknown nominal noise variances.
Unlike ASOR's covariance-scaling formulation, D-ASOR
employs an outlier likelihood that is independent of
the nominal noise variance.
Our contributions are threefold:
(i) a decoupled hierarchical observation model enabling
robust channel-wise nominal noise learning;
(ii) a variational expectation--maximization (VEM)
inference framework combining closed-form noise-parameter
and outlier probability updates with nonlinear Gaussian
Rauch--Tung--Striebel (RTS) smoothing; and
(iii) a theoretical characterization of the vanishing
direct influence of extreme outliers on nominal-variance
learning under fixed variational quantities and model
parameters, contrasting the proposed formulation with
a coupled covariance-learning counterpart of ASOR.
Numerical experiments evaluate estimation accuracy,
computational efficiency, and sensitivity to prior
specification.

\section{D-ASOR Methodology}
\label{sec:methodology}

\subsection{Decoupled Observation Model}
\label{subsec:observation_model}

Consider the discrete-time nonlinear state-space model
\begin{equation}
\begin{aligned}
\mathbf{x}_k &= f(\mathbf{x}_{k-1})+\mathbf{w}_{k-1},\\
\mathbf{y}_k &= h(\mathbf{x}_k)+\mathbf{e}_k,
\end{aligned}
\label{eq:state_space}
\end{equation}
where $\mathbf{x}_k\in\mathbb{R}^{n}$ and
$\mathbf{y}_k\in\mathbb{R}^{m}$ denote the state and
measurement vectors, respectively, $f$ and $h$
are known mappings, and $k$ is the time step.
The temporally independent process noise is modeled as
$\mathbf{w}_k\sim\mathcal{N}(\mathbf{0},\mathbf{Q}_k)$,
where $\mathcal{N}$ denotes the Gaussian distribution
with specified mean and covariance.
The vector $\mathbf{e}_k=[e_{k1},\ldots,e_{km}]^\top$
denotes the measurement noise vector.
Process and measurement noises are assumed mutually
independent. Measurement noise components are modeled
as conditionally independent across time and channels,
given the latent noise variables and parameters.

Let $s_i=\sigma_i^2>0$ denote the unknown nominal
measurement-noise variance in channel $i$,
assumed constant over time, and let
$\sigma_{0i}^2$ denote its prior estimate. We introduce an indicator
$\mathcal{I}_{ki}>0$, which equals one for nominal
observations and represents the outlier precision
otherwise. The decoupled observation likelihood is
\begin{equation}
\begin{aligned}
&p(y_{ki}\mid\mathbf{x}_k,\mathcal{I}_{ki},s_i)=\begin{cases}
\mathcal{N}(y_{ki};h_i(\mathbf{x}_k),s_i),
&\mathcal{I}_{ki}=1,\\
\mathcal{N}(y_{ki};h_i(\mathbf{x}_k),
\mathcal{I}_{ki}^{-1}),
&\mathcal{I}_{ki}\ne1.
\end{cases}
\end{aligned}
\label{eq:decoupled_likelihood}
\end{equation}
where $\mathcal{N}(y;\mu,v)$ denotes the Gaussian
density evaluated at $y$, with mean $\mu$ and
variance $v$, and $h_i$ is the $i$th component
of $h$. The value $\mathcal{I}_{ki}=1$ identifies
the nominal branch but does not set its precision;
the nominal precision remains $s_i^{-1}$.
Therefore, the conditional measurement-noise covariance
$\mathbf R_k$ is diagonal, with
$R_{k,ii}=s_i$ when $\mathcal I_{ki}=1$ and
$R_{k,ii}=\mathcal I_{ki}^{-1}$ otherwise.
Unlike covariance-scaling approaches
\cite{nakabayashi2020,chughtai2024heuristics,piche2012recursive},
the outlier variance $\mathcal{I}_{ki}^{-1}$ is
independent of $s_i$. Thus, the two likelihood
branches have decoupled variance parameters.

We assign the hierarchical priors
\begin{equation}
\begin{aligned}
s_i &\sim \operatorname{IG}(\alpha_{0i},\beta_{0i}),
\quad b_i \sim \operatorname{Ga}(A,B_{0i}),\\
p(\mathcal{I}_{ki}\mid b_i)
&=\theta\delta(\mathcal{I}_{ki}-1)
+(1-\theta)\operatorname{Ga}(\mathcal{I}_{ki};a,b_i),
\end{aligned}
\label{eq:observation_priors}
\end{equation}
where $\operatorname{IG}$ and $\operatorname{Ga}$ denote
the inverse-Gamma and Gamma distributions with
shape--scale and shape--rate parameters, respectively.
The inverse-Gamma and Gamma priors are chosen
for their conjugacy with variance- and precision-based
likelihoods, respectively, enabling closed-form updates.
The parameter $\theta\in(0,1)$ is the prior nominal
probability, and $\delta(\cdot)$ denotes the Dirac
delta function.
The point mass selects the nominal branch, while
the Gamma component describes the outlier precision.
The inverse-Gamma and Gamma densities are
\begin{equation*}
\begin{aligned}
\operatorname{IG}(s;\alpha,\beta)
&=\frac{\beta^\alpha}{\Gamma(\alpha)}
s^{-\alpha-1}e^{-\beta/s},\\
\operatorname{Ga}(u;a,b)
&=\frac{b^a}{\Gamma(a)}u^{a-1}e^{-bu},
\end{aligned}
\end{equation*}
for $s,u>0$, where $\Gamma(\cdot)$ is the
Gamma function. Each channel has its own nominal
variance $s_i$ and Gamma rate parameter $b_i$,
both shared across time, with mutually independent
priors. The indicators $\mathcal{I}_{ki}$
are observation-specific and conditionally
independent across observations given $\mathbf b$.
The nominal probability $\theta$ and
hyperparameters $a>0$, $A>1$, and $B_{0i}>0$
are fixed. Each $b_i$ is estimated using
observations of its corresponding channel.

The resulting conditional measurement likelihood
factorizes across channels as
\begin{equation*}
\begin{aligned}
&p(\mathbf{y}_k\mid\mathbf{x}_k,
\boldsymbol{\mathcal I}_k,\mathbf{s})=\prod_{i=1}^{m}
p(y_{ki}\mid\mathbf{x}_k,\mathcal{I}_{ki},s_i).
\end{aligned}
\end{equation*}

\subsection{Variational EM-Based Smoothing}
\label{subsec:variational_inference}

Since, exact Bayesian inference is intractable we adopt a
variational expectation--maximization (VEM) framework. We use approximate Bayesian inference for the latent
variables and maximum a posteriori (MAP) estimation of the channel-specific
Gamma rate parameters $b_i$, following ASOR
\cite{chughtai2024heuristics}. This simplifies
variational updates while preserving closed-form
inference for the noise-related latent variables.

We use $\mathbf z_{j:k}=(\mathbf z_j,\ldots,\mathbf z_k)$
to denote a sequence from time $j$ to $k$. Let $Y=\mathbf{y}_{2:K}$,
$Z=(\mathbf{x}_{1:K},\mathbf{s},
\boldsymbol{\mathcal I}_{2:K})$,
and $\mathbf{b}=(b_1,\ldots,b_m)$, where
$\mathbf{s}=(s_1,\ldots,s_m)$ and
$\boldsymbol{\mathcal I}_k=
(\mathcal I_{k1},\ldots,\mathcal I_{km})$.
Assuming $\mathbf{x}_1\sim
\mathcal{N}(\mathbf{m}_1,\mathbf{P}_1)$,
the trajectory prior is
\begin{equation*}
p(\mathbf{x}_{1:K})=
\mathcal{N}(\mathbf{x}_1;\mathbf{m}_1,\mathbf{P}_1)
\prod_{k=2}^{K}
\mathcal{N}\!\left(
\mathbf{x}_k;f(\mathbf{x}_{k-1}),\mathbf{Q}_{k-1}
\right).
\end{equation*}

The complete joint distribution
$p(Y,Z,\mathbf b)$ factorizes as
\begin{equation*}
\begin{aligned}
&p(\mathbf{x}_{1:K})\prod_{i=1}^{m}
\Bigg[p(s_i)p(b_i)
\prod_{k=2}^{K}
p(\mathcal I_{ki}\mid b_i)
p(y_{ki}\mid\mathbf{x}_k,\mathcal I_{ki},s_i)
\Bigg].
\end{aligned}
\end{equation*}

\noindent\textit{Variational E-step:}
For fixed $\mathbf b$, we approximate the posterior by
\begin{equation}
\begin{aligned}
p(Z\mid Y,\mathbf b)\approx q(Z)
&=q(\mathbf{x}_{1:K})\prod_{i=1}^{m}q(s_i)\prod_{k=2}^{K}\prod_{i=1}^{m}
q(\mathcal{I}_{ki}).
\end{aligned}
\label{eq:variational_factorization}
\end{equation}

This factorization retains temporal dependence
within the trajectory. Throughout the derivation,
$C$ denotes an additive constant independent of
the variable being updated. Variational coordinate
updates take the form
\begin{equation}
\log q^\star(\Delta)
=\mathbb{E}_{q_{-\Delta}}
\!\left[\log p(Y,Z,\mathbf b)\right]
+C,
\label{eq:coordinate_update}
\end{equation}
where $\Delta$ is the updated variable block
and $q_{-\Delta}$ denotes the remaining factors.

\noindent\textit{State update:}
Let $\Omega_{ki}=q(\mathcal{I}_{ki}=1)$ denote
the nominal posterior probability, and define
the conditional mean outlier precision
$\bar{\mathcal I}_{ki}=
\mathbb E_q[\mathcal I_{ki}\mid\mathcal I_{ki}\ne1]$.
Applying \eqref{eq:coordinate_update} to the
trajectory gives
\begin{equation*}
\begin{aligned}
\log q^\star(\mathbf{x}_{1:K})=\log p(\mathbf{x}_{1:K})-\frac12\sum_{k=2}^{K}\sum_{i=1}^{m}
\tau_{ki}[y_{ki}-h_i(\mathbf{x}_k)]^2+C,
\end{aligned}
\end{equation*}
where the effective measurement precision is
\begin{equation}
\tau_{ki}=
\Omega_{ki}\mathbb E_{q(s_i)}[s_i^{-1}]
+(1-\Omega_{ki})\bar{\mathcal I}_{ki}.
\label{eq:effective_precision}
\end{equation}

The state update uses the diagonal effective covariance
$\mathbf T_k^{-1}$ with $\mathbf T_k=\operatorname{diag}(\tau_{k1},\ldots,\tau_{km})$.
For fixed $\mathbf T_k$, nonlinear Gaussian smoothing
approximates the state marginals as
$q(\mathbf x_k)\approx\mathcal N(\mathbf x_k;
\mathbf m_k^s,\mathbf P_k^s)$ \cite{sarkka2010gaussian}.
Let
$q_k^f(\mathbf x)=\mathcal N(\mathbf x;\mathbf m_k^f,\mathbf P_k^f)$
and
$q_k^-(\mathbf x)=\mathcal N(\mathbf x;\mathbf m_k^-,\mathbf P_k^-)$
denote the filtered and predicted Gaussian approximations.

\noindent\textit{Forward pass:}\enspace
Starting from $(\mathbf m_1^f,\mathbf P_1^f)
=(\mathbf m_1,\mathbf P_1)$, the prediction moments
for $k=2,\ldots,K$ are
\begin{equation}
\begin{aligned}
\mathbf m_k^-&=\int f(\mathbf x)q_{k-1}^f(\mathbf x)
\,d\mathbf x,\\
\mathbf P_k^-&=\mathbf Q_{k-1}
+\int\tilde f_k(\mathbf x)\tilde f_k(\mathbf x)^\top
q_{k-1}^f(\mathbf x)\,d\mathbf x,
\end{aligned}
\label{eq:gaussian_prediction}
\end{equation}
where $\tilde f_k(\mathbf x)=f(\mathbf x)-\mathbf m_k^-$. The measurement moments are
\begin{equation}
\begin{aligned}
\bar{\mathbf y}_k&=\int h(\mathbf x)q_k^-(\mathbf x)
\,d\mathbf x,\\
\mathbf S_k&=\mathbf T_k^{-1}
+\int\tilde h_k(\mathbf x)\tilde h_k(\mathbf x)^\top
q_k^-(\mathbf x)\,d\mathbf x,\\
\mathbf U_k&=\int(\mathbf x-\mathbf m_k^-)
\tilde h_k(\mathbf x)^\top q_k^-(\mathbf x)\,d\mathbf x,
\end{aligned}
\label{eq:gaussian_measurement_moments}
\end{equation}
where $\tilde h_k(\mathbf x)=h(\mathbf x)-\bar{\mathbf y}_k$.
Gaussian conditioning gives
\begin{equation}
\begin{aligned}
\mathbf K_k&=\mathbf U_k\mathbf S_k^{-1},\quad
\mathbf m_k^f=\mathbf m_k^-+
\mathbf K_k(\mathbf y_k-\bar{\mathbf y}_k),\\
\mathbf P_k^f&=\mathbf P_k^--
\mathbf K_k\mathbf S_k\mathbf K_k^\top.
\end{aligned}
\label{eq:gaussian_filter_update}
\end{equation}

\noindent\textit{Backward pass:}\enspace
Define the transition cross-covariance
\begin{equation}
\mathbf C_k=\int(\mathbf x-\mathbf m_k^f)
\tilde f_{k+1}(\mathbf x)^\top q_k^f(\mathbf x)
\,d\mathbf x.
\label{eq:gaussian_transition_cross_covariance}
\end{equation}

Initialize $(\mathbf m_K^s,\mathbf P_K^s)
=(\mathbf m_K^f,\mathbf P_K^f)$. For
$k=K-1,\ldots,1$, apply
\begin{equation}
\begin{aligned}
\mathbf G_k&=\mathbf C_k(\mathbf P_{k+1}^-)^{-1},\quad
\mathbf m_k^s=\mathbf m_k^f+
\mathbf G_k(\mathbf m_{k+1}^s-\mathbf m_{k+1}^-),\\
\mathbf P_k^s&=\mathbf P_k^f+
\mathbf G_k(\mathbf P_{k+1}^s-\mathbf P_{k+1}^-)
\mathbf G_k^\top.
\end{aligned}
\label{eq:gaussian_rts_smoothing}
\end{equation}

The remaining updates use the expected squared residual
\begin{equation}
\begin{aligned}
D_{ki}&=\mathbb E_{q(\mathbf{x}_k)}
[(y_{ki}-h_i(\mathbf{x}_k))^2]=(y_{ki}-\bar h_{ki}^s)^2+V_{ki}^s,
\end{aligned}
\label{eq:expected_residual}
\end{equation}
where $\bar h_{ki}^s=
\mathbb E_{q(\mathbf{x}_k)}[h_i(\mathbf{x}_k)]$ and
$V_{ki}^s=\operatorname{Var}_{q(\mathbf{x}_k)}
[h_i(\mathbf{x}_k)]$ are evaluated using the
chosen Gaussian moment approximation.
The variance term accounts for uncertainty
in the smoothed state estimate.

\noindent\textit{Mixed-indicator update:}
Applying \eqref{eq:coordinate_update} to
$\mathcal I_{ki}$ yields a spike-and-Gamma posterior
\begin{equation*}
\begin{aligned}
q(\mathcal I_{ki})
&=\Omega_{ki}\delta(\mathcal I_{ki}-1)\\
&\quad+(1-\Omega_{ki})\operatorname{Ga}\!\left(
\mathcal I_{ki};a+0.5,
 b_i+0.5{D_{ki}}\right).
\end{aligned}
\end{equation*}
The unnormalized nominal and outlier weights are
\begin{equation}
\begin{aligned}
Z_{ki}^{\mathrm n}
&=\theta\exp\!\left[-\frac12\left(
\mathbb E_{q(s_i)}[\log s_i]
+D_{ki}\mathbb E_{q(s_i)}[s_i^{-1}]
\right)\right],\\
Z_{ki}^{\mathrm o}
&=(1-\theta)
\frac{b_i^a\Gamma(a+1/2)}
{\Gamma(a)(b_i+D_{ki}/2)^{a+1/2}}.
\end{aligned}
\label{eq:branch_weights}
\end{equation}

The posterior nominal probability and conditional
mean outlier precision are then
\begin{equation}
\Omega_{ki}=\frac{Z_{ki}^{\mathrm n}}
{Z_{ki}^{\mathrm n}+Z_{ki}^{\mathrm o}},
\qquad
\bar{\mathcal I}_{ki}=
\frac{a+1/2}{b_i+D_{ki}/2},
\label{eq:indicator_precision_update}
\end{equation}
which determine the effective measurement precision
in \eqref{eq:effective_precision}.

\begin{algorithm}[t!]
\caption{The proposed smoother: D-ASOR}
\label{alg:dasor}

Initialize $\alpha_i=\alpha_{0i}$,
$\beta_i=\beta_{0i}$,
$b_i=b_i^{(0)}$,
$\tau_{ki}=\sigma_{0i}^{-2}$ for all $i$\;

\While{stopping criterion not met}{

    Initialize $(\mathbf m_1^f,\mathbf P_1^f)
    =(\mathbf m_1,\mathbf P_1)$\;

    \For{$k=2,\ldots,K$}{
        Evaluate $(\mathbf m_k^-,\mathbf P_k^-)$
        using \eqref{eq:gaussian_prediction}\;

        Evaluate $(\mathbf m_k^f,\mathbf P_k^f)$
        using \eqref{eq:gaussian_measurement_moments}
        --\eqref{eq:gaussian_filter_update}\;
    }

    $(\mathbf m_K^s,\mathbf P_K^s)
    =(\mathbf m_K^f,\mathbf P_K^f)$\;

    \For{$k=K-1,\ldots,1$}{
        Evaluate $(\mathbf m_k^s,\mathbf P_k^s)$
        using
        \eqref{eq:gaussian_transition_cross_covariance}
        --\eqref{eq:gaussian_rts_smoothing}\;
    }

    \For{$k=2,\ldots,K$}{
        Evaluate $D_{ki}$ and update
        $(\Omega_{ki},\bar{\mathcal I}_{ki})$
        for all $i$ using
        \eqref{eq:expected_residual},
        \eqref{eq:branch_weights}, and
        \eqref{eq:indicator_precision_update}\;
    }

    \For{$i=1,\ldots,m$}{
        Update $(\alpha_i,\beta_i)$ and $b_i$
        using \eqref{eq:nominal_variance_posterior}
        and \eqref{eq:rate_update}\;
    }

    Update $\tau_{ki}$ for all $k,i$
    using \eqref{eq:effective_precision}\;
}

Repeat forward and backward passes
with final $\tau_{ki}$\;

\end{algorithm}

\begin{table*}[!t]
\centering
\caption{State NRMSE $E_x$ over 30 paired MC trials.
Bold indicates the lowest non-oracle error.}
\label{tab:state_sigma}

\begingroup
\fontsize{9}{10}\selectfont
\setlength{\tabcolsep}{3pt}
\renewcommand{\arraystretch}{1.0}

\begin{tabular*}{\textwidth}
{@{\extracolsep{\fill}}cc*{10}{r}@{}}
\toprule
& & \multicolumn{5}{c}{Gaussian contamination}
& \multicolumn{5}{c}{Uniform contamination}\\
\cmidrule(lr){3-7}
\cmidrule(lr){8-12}

$\rho$ & $p_{\rm out}$
& IDEAL & D-ASOR & VB-H & GGM & UKF
& IDEAL & D-ASOR & VB-H & GGM & UKF\\
\midrule

5 & 0.2
& 0.0383 & \textbf{0.0393} & 0.0407 & 0.0434 & 0.0554
& 0.0383 & \textbf{0.0390} & 0.0422 & 0.0503 & 0.0839\\

& 0.4
& 0.0372 & \textbf{0.0394} & 0.0421 & 0.0463 & 0.0741
& 0.0370 & \textbf{0.0390} & 0.0515 & 0.0644 & 0.1113\\

& 0.6
& 0.0425 & \textbf{0.0477} & 0.0500 & 0.0555 & 0.0799
& 0.0425 & \textbf{0.0560} & 0.0703 & 0.0761 & 0.6663\\

\midrule

10 & 0.2
& 0.0349 & \textbf{0.0358} & 0.0381 & 0.0439 & 0.0806
& 0.0349 & \textbf{0.0354} & 0.0398 & 0.0643 & 1.1474\\

& 0.4
& 0.0374 & \textbf{0.0395} & 0.0490 & 0.0638 & 0.2990
& 0.0374 & \textbf{0.0387} & 0.0647 & 0.1478 & 44.7001\\

& 0.6
& 0.0435 & \textbf{0.0476} & 0.0608 & 0.0693 & 7.7045
& 0.0434 & \textbf{0.0812} & 0.1207 & 0.1305 & 35.6584\\

\bottomrule
\end{tabular*}

\par\vspace{2pt}
{\footnotesize
IDEAL: oracle UKF--RTS;
VB-H: VB-HAUKF;
GGM: projected-diagonal VBAKF-GGM;
UKF: fixed-covariance UKF--RTS.\par}

\endgroup
\end{table*}

\begin{figure*}[!t]
\centering
\begin{minipage}[b]{0.32\textwidth}
\centering
\includegraphics[width=\linewidth,height=3.2cm,
keepaspectratio]{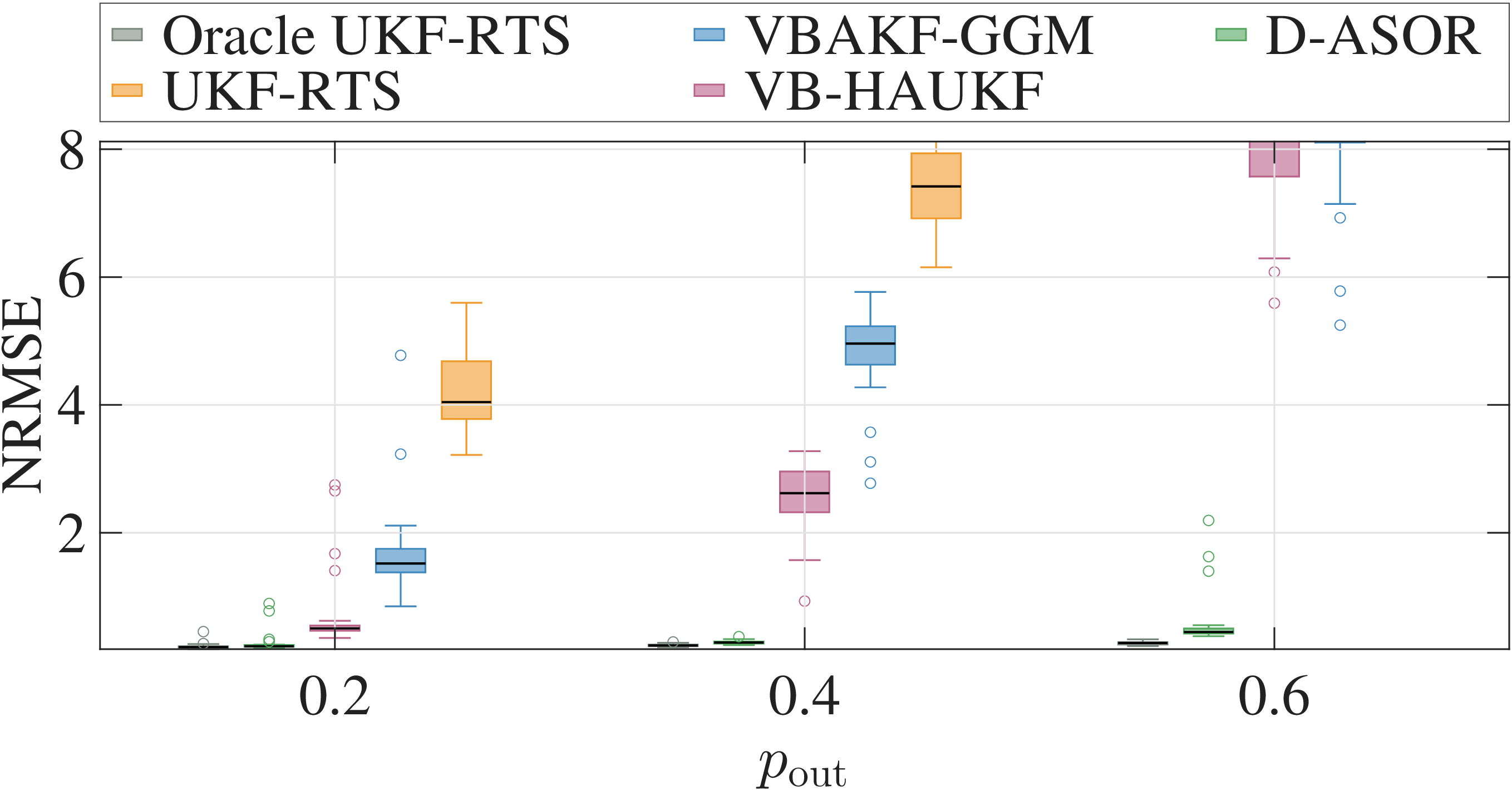}
\par\vspace{2pt}
{\footnotesize (a) Uniform contamination\par}
\end{minipage}\hfill
\begin{minipage}[b]{0.32\textwidth}
\centering
\includegraphics[width=\linewidth,height=3.2cm,
keepaspectratio]{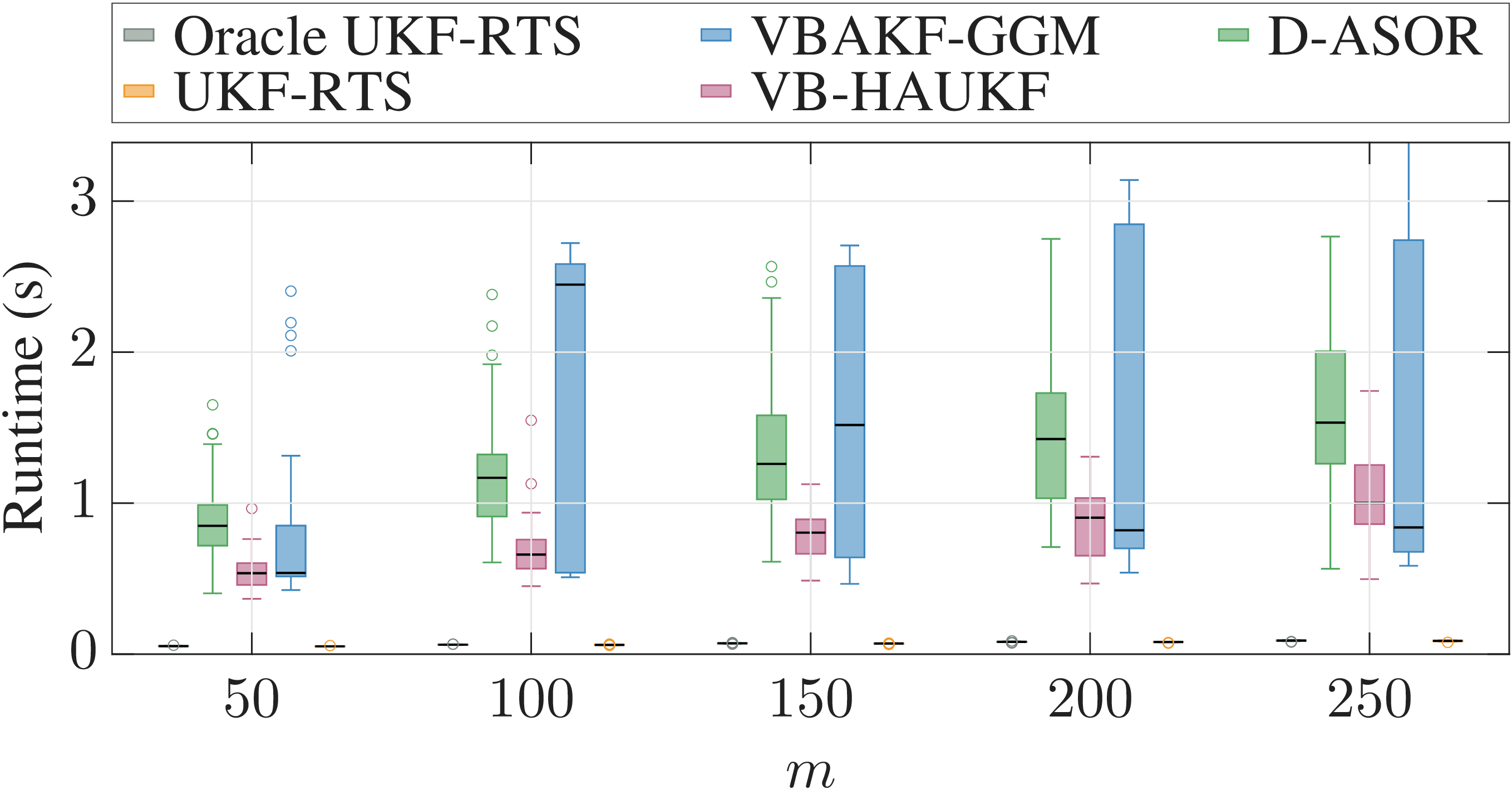}
\par\vspace{2pt}
{\footnotesize (b) Computational cost\par}
\end{minipage}\hfill
\begin{minipage}[b]{0.32\textwidth}
\centering
\includegraphics[width=\linewidth,height=3.2cm,
keepaspectratio]{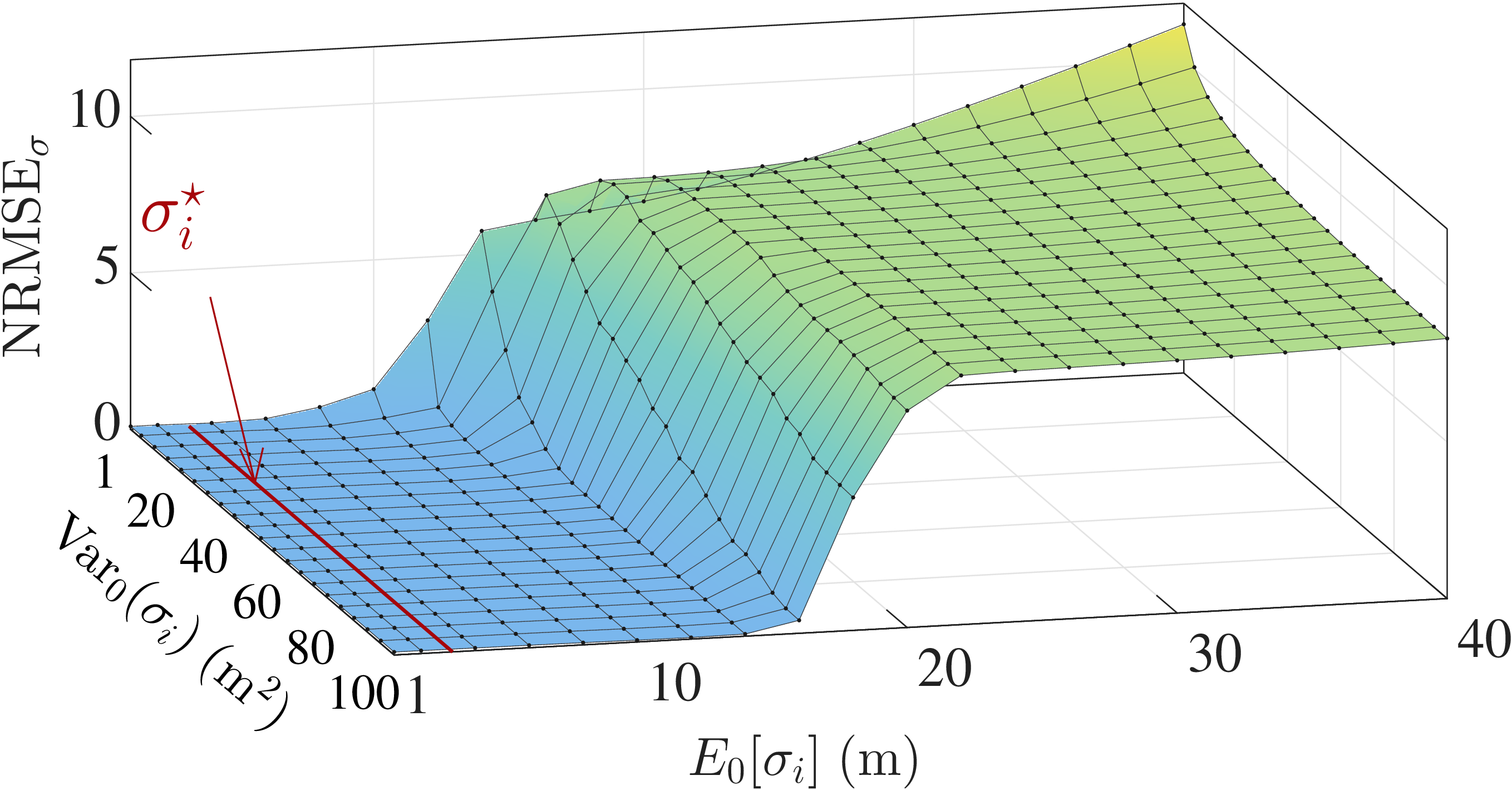}
\par\vspace{2pt}
{\footnotesize (c) Prior sensitivity\par}
\end{minipage}

\caption{Numerical results:
(a) measurement-reconstruction NRMSE under uniform
contamination at $\rho=10$;
(b) runtime versus measurement dimension $m$;
(c) nominal-noise prior sensitivity under Gaussian
contamination at $(\rho,p_{\rm out})=(10,0.4)$.}
\label{fig:main_results}
\end{figure*}

\noindent\textit{Nominal-variance update:}
Only the nominal likelihood branch depends on $s_i$.
Applying \eqref{eq:coordinate_update} gives
\begin{equation*}
\begin{aligned}
\log q^\star(s_i)
&=\log p(s_i)-\frac12\sum_{k=2}^{K}\Omega_{ki}
\left(\log s_i+\frac{D_{ki}}{s_i}\right)
+C.
\end{aligned}
\end{equation*}
Combining these terms with the inverse-Gamma prior yields
\begin{equation}
\begin{gathered}
q(s_i)=\operatorname{IG}(s_i;\alpha_i,\beta_i),\\
\alpha_i=\alpha_{0i}+\frac12\sum_{k=2}^{K}\Omega_{ki},
\quad \beta_i=\beta_{0i}+\frac12\sum_{k=2}^{K}
\Omega_{ki}D_{ki}.
\end{gathered}
\label{eq:nominal_variance_posterior}
\end{equation}

Posterior nominal probabilities limit outlier influence
on variance learning. The required moments are
\begin{equation*}
\begin{aligned}
\mathbb E_{q(s_i)}[s_i^{-1}]
&=\frac{\alpha_i}{\beta_i}, \ 
\mathbb E_{q(s_i)}[\log s_i]
=\log\beta_i-\psi(\alpha_i),
\end{aligned}
\end{equation*}
where $\psi$ is the digamma function.

\noindent\textit{MAP M-step:}
For fixed variational factors, each $b_i$ maximizes the
expected log joint, including its Gamma prior. The relevant
terms are
\begin{equation*}
\begin{aligned}
\mathcal J_i(b_i)
&=(A-1)\log b_i-B_{0i}b_i\\
&\quad+\sum_{k=2}^{K}(1-\Omega_{ki})
\left(a\log b_i-b_i\bar{\mathcal I}_{ki}\right)
+C.
\end{aligned}
\end{equation*}

Setting $\partial\mathcal J_i/\partial b_i=0$ gives
\begin{equation}
b_i^+={
(A-1+a\displaystyle\sum_{k=2}^{K}(1-\Omega_{ki}))
}/{
(B_{0i}+\displaystyle\sum_{k=2}^{K}
(1-\Omega_{ki})\bar{\mathcal I}_{ki})
}.
\label{eq:rate_update}
\end{equation}

For $A>1$ and $B_{0i}>0$, this is the unique positive
maximizer, used in the subsequent variational E-step.

The outer iterations terminate based on the relative
$\ell_2$-norm change between successive stacked
smoothed state means \cite{nakabayashi2020,10679915}. Algorithm~\ref{alg:dasor} summarizes D-ASOR.

\subsection{Robustness of Nominal-Variance Learning}
\label{subsec:robustness}

For fixed variational moments of $s_i$ and fixed
outlier-model parameters, the posterior nominal
probability satisfies
\begin{equation}
\lim_{D_{ki}\to\infty}\Omega_{ki}D_{ki}=0.
\label{eq:vanishing_influence}
\end{equation}

Consequently, increasingly extreme residuals have
vanishing direct influence on nominal-variance learning.
The proof, comparison with a coupled covariance-learning
counterpart of ASOR, and numerical validation are provided
in the \emph{Supplementary Material} document.

\section{Numerical Experiments}
\label{sec:numerical_experiments}

We evaluate D-ASOR in nonlinear target tracking with five
range and five bearing measurements, comparing it with
fixed-covariance UKF--RTS, VB-HAUKF \cite{li2016vbhaukf},
projected-diagonal VBAKF-GGM \cite{zhu2022},
and oracle UKF--RTS, which uses known contamination
statistics as a reference. For a matched comparison, VB-HAUKF
and VBAKF-GGM are implemented within an iterative RTS
smoothing framework. Gaussian moment integrals use the
unscented transform \cite{sarkka2008}, with wrapped bearing differences.
To assess robustness to different outlier statistics,
we consider Gaussian $\mathcal N(0,\rho^2\sigma_i^2)$
and uniform $\mathcal U(0,\sqrt{12}\rho\sigma_i)$
contamination. Both distributions have standard deviation
$\rho\sigma_i$, while the latter has nonzero mean.
We test for different values of $\rho$ and outlier probabilities $p_{\rm out}$ over 30 paired Monte Carlo (MC)
trials per configuration. 
All iterative adaptive methods use the same
convergence criterion as D-ASOR. Model parameters, evaluation metrics, and implementation
details are provided in the \emph{Information for
Reproducibility} document.

Table~\ref{tab:state_sigma} reports state normalized
root-mean-square error (NRMSE) for $\rho=5$ and $10$.
D-ASOR achieves the lowest non-oracle error in all reported configurations,
often approaching the oracle reference.
The improvements are particularly pronounced under
uniform contamination: at
$(\rho,p_{\rm out})=(10,0.4)$, D-ASOR reduces the
state NRMSE by approximately $40\%$ and $74\%$
relative to VB-HAUKF and VBAKF-GGM, respectively.
Under uniform contamination at $p_{\rm out}=0.6$, the gap
from the oracle widens as contaminated observations predominate.
Similar trends are generally observed for $\rho=15$ and $20$.

Beyond state estimation, Fig.~\ref{fig:main_results}(a)
examines noise-free measurement reconstruction under
uniform contamination at $\rho=10$.
D-ASOR achieves the lowest median reconstruction NRMSE
among the non-oracle methods at all tested outlier
probabilities, although accuracy deteriorates as
$p_{\rm out}$ increases. Similar trends are observed
under Gaussian contamination.

Regarding computational cost, D-ASOR jointly processes
all $m$ measurement channels at each time step using diagonal
measurement precision and a fixed-size solve in sigma-point space,
resulting in $\mathcal{O}(m)$ complexity per iteration for fixed
$n$ and $K$. Figure~\ref{fig:main_results}(b) compares runtimes,
with VB-HAUKF being the fastest adaptive smoother, 
followed generally by VBAKF-GGM and D-ASOR. The computational overhead of D-ASOR is accompanied
by improved estimation accuracy and robustness. 

Finally, Fig.~\ref{fig:main_results}(c) examines the
sensitivity of nominal-noise estimation to the prior mean
and variance of $\sigma_i$ for range channels with true
standard deviation $\sigma_i^\star=\sqrt{10}\,\mathrm{m}$,
marked in red.
Increasing the prior variance extends the range of prior
means over which accurate noise estimation is achieved
by allowing the observations to exert greater influence
on nominal-noise learning. However, specifying prior
means far above the true value remains challenging
within the tested range of prior variances.

\section{Conclusion}
\label{sec:conclusion}

The proposed D-ASOR smoother jointly estimates nonlinear
state trajectories and channel-specific nominal noise variances
in the presence of outliers. 
Its decoupled observation model yields vanishing
direct influence of extreme outliers on
nominal-variance learning under fixed variational
quantities and model parameters. Numerical experiments demonstrate improved state estimation
and measurement reconstruction under Gaussian and uniform
contamination. For fixed state dimension and trajectory length, D-ASOR's joint
update scales linearly with the number of measurement channels
per iteration.

\clearpage
\bibliographystyle{IEEEtran}
\bibliography{references}

@IEEEtranBSTCTL{BSTcontrol,
  CTLdash_repeated_names = "no"
}

@article{sarkka2008,
  author  = {S{\"a}rkk{\"a}, Simo},
  title   = {Unscented {Rauch--Tung--Striebel} Smoother},
  journal = {IEEE Transactions on Automatic Control},
  volume  = {53},
  number  = {3},
  pages   = {845--849},
  year    = {2008}
}

@article{majal2024,
  author  = {Majal, Arslan and Chughtai, Aamir Hussain},
  title   = {Outlier-Robust Unscented {RTS} Smoothing
             for Independent Sensing Data},
  journal = {IEEE Sensors Letters},
  volume  = {8},
  number  = {10},
  pages   = {1--4},
  year    = {2024},
  doi     = {10.1109/LSENS.2024.3460975}
}

@article{zhu2022,
  author  = {Zhu, Hao and Zhang, Guorui
             and Li, Yongfu and Leung, Henry},
  title   = {An Adaptive {Kalman} Filter With Inaccurate Noise
             Covariances in the Presence of Outliers},
  journal = {IEEE Transactions on Automatic Control},
  volume  = {67},
  number  = {1},
  pages   = {374--381},
  year    = {2022},
  doi     = {10.1109/TAC.2021.3056343}
}

@article{sarkka2010gaussian,
  author  = {S{\"a}rkk{\"a}, Simo and Hartikainen, Jouni},
  title   = {On {Gaussian} Optimal Smoothing of Non-Linear
             State Space Models},
  journal = {IEEE Transactions on Automatic Control},
  volume  = {55},
  number  = {8},
  pages   = {1938--1941},
  year    = {2010}
}

@article{wang2018robust,
  author  = {Wang, Hongwei and Li, Hongbin
             and Fang, Jun and Wang, Heping},
  title   = {Robust {Gaussian} {Kalman} Filter
             With Outlier Detection},
  journal = {IEEE Signal Processing Letters},
  volume  = {25},
  number  = {8},
  pages   = {1236--1240},
  year    = {2018},
  publisher = {IEEE}
}

@article{chughtai2022,
  author  = {Chughtai, Aamir Hussain and Tahir, Muhammad
             and Uppal, Momin},
  title   = {Outlier-Robust Filtering for Nonlinear Systems
             With Selective Observations Rejection},
  journal = {IEEE Sensors Journal},
  volume  = {22},
  number  = {7},
  pages   = {6887--6897},
  year    = {2022},
  doi     = {10.1109/JSEN.2022.3152609}
}

@article{li2016vbhaukf,
  author  = {Li, Kailong and Chang, Lubin and Hu, Baiqing},
  title   = {A Variational {Bayesian}-Based Unscented {Kalman}
             Filter With Both Adaptivity and Robustness},
  journal = {IEEE Sensors Journal},
  volume  = {16},
  number  = {18},
  pages   = {6966--6976},
  year    = {2016},
  doi     = {10.1109/JSEN.2016.2591260}
}

@article{chughtai2024heuristics,
  author  = {Chughtai, Aamir Hussain and Tahir, Muhammad
             and Uppal, Momin},
  title   = {{Bayesian} Heuristics for Robust Spatial Perception},
  journal = {IEEE Transactions on Instrumentation and Measurement},
  volume  = {73},
  pages   = {1--11},
  year    = {2024},
  doi     = {10.1109/TIM.2024.3364269}
}

@article{nakabayashi2020,
  author  = {Nakabayashi, Akio and Ueno, Genta},
  title   = {Nonlinear Filtering Method Using a Switching
             Error Model for Outlier-Contaminated Observations},
  journal = {IEEE Transactions on Automatic Control},
  volume  = {65},
  number  = {7},
  pages   = {3150--3156},
  year    = {2020},
  doi     = {10.1109/TAC.2019.2947649}
}

@inproceedings{majal2025emorf,
  author       = {Majal, Arslan and Chughtai, Aamir Hussain
                  and Tahir, Muhammad},
  title        = {{EMORF-II}: Adaptive {EM}-Based Outlier-Robust
                  Filtering With Correlated Measurement Noise},
  booktitle    = {2025 {IEEE} 35th International Workshop
                  on Machine Learning for Signal Processing ({MLSP})},
  pages        = {1--6},
  year         = {2025},
  organization = {IEEE}
}

@article{10679915,
  author  = {Chughtai, Aamir Hussain and Tahir, Muhammad
             and Uppal, Momin},
  title   = {{EMORF/S}: {EM}-Based Outlier-Robust Filtering
             and Smoothing With Correlated Measurement Noise},
  journal = {IEEE Transactions on Signal Processing},
  volume  = {72},
  pages   = {4318--4331},
  year    = {2024},
  doi     = {10.1109/TSP.2024.3460176}
}

@article{huang2016smoother,
  author  = {Huang, Yulong and Zhang, Yonggang and Li, Ning
             and Chambers, Jonathon A.},
  title   = {A Robust {Gaussian} Approximate Fixed-Interval
             Smoother for Nonlinear Systems With Heavy-Tailed
             Process and Measurement Noises},
  journal = {IEEE Signal Processing Letters},
  volume  = {23},
  number  = {4},
  pages   = {468--472},
  year    = {2016},
  doi     = {10.1109/LSP.2016.2533543}
}

@article{chen2020smoother,
  author  = {Chen, Yuming and Li, Wei and Wang, Yuqiao},
  title   = {Robust {Gaussian} Approximate Fixed-Interval
             Smoother With Outlier Detection},
  journal = {IEEE Signal Processing Letters},
  volume  = {27},
  pages   = {1505--1509},
  year    = {2020},
  doi     = {10.1109/LSP.2020.3015422}
}

@article{bai2021,
  author  = {Bai, Mingming and Huang, Yulong
             and Jia, Guangle and Zhang, Yonggang},
  title   = {A Robust Fixed-Interval Smoother for Nonlinear
             Systems With Non-Stationary Heavy-Tailed
             State and Measurement Noises},
  journal = {Signal Processing},
  volume  = {180},
  pages   = {107898},
  year    = {2021},
  doi     = {10.1016/j.sigpro.2020.107898}
}

@article{aravkin2014,
  author  = {Aravkin, Aleksandr Y. and Burke, James V.
             and Pillonetto, Gianluigi},
  title   = {Robust and Trend-Following {Student's} {$t$}
             {Kalman} Smoothers},
  journal = {SIAM Journal on Control and Optimization},
  volume  = {52},
  number  = {5},
  pages   = {2891--2916},
  year    = {2014},
  doi     = {10.1137/130918861}
}

@article{li2021robust,
  author  = {Li, Haoqing and Medina, Daniel
             and Vil{\`a}-Valls, Jordi and Closas, Pau},
  title   = {Robust Variational-Based {Kalman} Filter
             for Outlier Rejection With Correlated
             Measurements},
  journal = {IEEE Transactions on Signal Processing},
  volume  = {69},
  pages   = {357--369},
  year    = {2021},
  doi     = {10.1109/TSP.2020.3042944}
}

@article{sarkka2009recursive,
  author  = {S{\"a}rkk{\"a}, Simo and Nummenmaa, Aapo},
  title   = {Recursive Noise Adaptive {Kalman} Filtering
             by Variational {Bayesian} Approximations},
  journal = {IEEE Transactions on Automatic Control},
  volume  = {54},
  number  = {3},
  pages   = {596--600},
  year    = {2009},
  doi     = {10.1109/TAC.2008.2008348}
}

@article{blei2017,
  author  = {Blei, David M. and Kucukelbir, Alp
             and McAuliffe, Jon D.},
  title   = {Variational Inference: A Review for Statisticians},
  journal = {Journal of the American Statistical Association},
  volume  = {112},
  number  = {518},
  pages   = {859--877},
  year    = {2017},
  doi     = {10.1080/01621459.2017.1285773}
}

@book{sarkka2023bayesian,
  author    = {S{\"a}rkk{\"a}, Simo and Svensson, Lennart},
  title     = {{Bayesian} Filtering and Smoothing},
  edition   = {2nd},
  year      = {2023},
  publisher = {Cambridge University Press}
}

@inproceedings{piche2012recursive,
  title={Recursive outlier-robust filtering and smoothing for nonlinear systems using the multivariate Student-t distribution},
  author={Pich{\'e}, Robert and S{\"a}rkk{\"a}, Simo and Hartikainen, Jouni},
  booktitle={2012 IEEE International Workshop on Machine Learning for Signal Processing},
  pages={1--6},
  year={2012},
  organization={IEEE}
}

\clearpage
\setlength{\abovedisplayskip}{3pt plus 1pt minus 1pt}
\setlength{\belowdisplayskip}{3pt plus 1pt minus 1pt}
\setlength{\abovedisplayshortskip}{2pt plus 1pt minus 1pt}
\setlength{\belowdisplayshortskip}{3pt plus 1pt minus 1pt}
\setlength{\jot}{1pt}

\twocolumn[{\begin{center}
{\bfseries Supplementary Material}\\[-1pt]
{\small Outlier-Robust Variational Bayesian Smoothing with Decoupled Hierarchical Likelihoods}
\end{center}}]

\section{Vanishing Outlier Influence}
\label{sec:supp_theory}

In D-ASOR, the nominal variance $s_i>0$ of channel $i$
has a variational posterior
$q(s_i)=\operatorname{IG}(s_i;\alpha_i,\beta_i)$,
with parameters
\begin{equation}
\begin{aligned}
\alpha_i&=\alpha_{0i}+\tfrac12\sum_{k=2}^K\Omega_{ki},\beta_i=\beta_{0i}+\tfrac12\sum_{k=2}^K\Omega_{ki}D_{ki}.
\end{aligned}
\label{eq:supp_variance_update}
\end{equation}
Here $D_{ki}=\mathbb E_{q(\mathbf x_k)}
[(y_{ki}-h_i(\mathbf x_k))^2]$ is the expected squared
residual and $\Omega_{ki}$ is the posterior nominal
probability. Since the outlier measurement variance is
independent of $s_i$, the influence of each observation
on nominal-variance learning is governed by $\Omega_{ki}$
and $\Omega_{ki}D_{ki}$. To examine this dependence,
we explicitly write the nominal probability as
$\Omega_{ki}(D)$.

\noindent\textit{Proposition 1 (Vanishing outlier influence):}
Fix a channel $i$ and let $D\geq0$ denote the expected
squared residual of one observation. For fixed
$u_i=\mathbb E_{q(s_i)}[s_i^{-1}]>0$,
$v_i=\mathbb E_{q(s_i)}[\log s_i]$, outlier rate
$b_i>0$, $a>0$, and $\theta\in(0,1)$,
\begin{equation}
\lim_{D\to\infty}D\Omega_{ki}(D)=0.
\label{eq:supp_vanishing}
\end{equation}
Consequently, the observation's contributions to both
inverse-Gamma parameters in \eqref{eq:supp_variance_update}
vanish for increasingly extreme residuals.

\noindent\textit{Proof:}
The unnormalized nominal and outlier branch weights are
\begin{equation}
\begin{aligned}
Z_{ki}^{\mathrm n}(D)
&=\theta\exp[-(v_i+u_iD)/2],\\
Z_{ki}^{\mathrm o}(D)
&=(1-\theta)\frac{b_i^a\Gamma(a+1/2)}
{\Gamma(a)(b_i+D/2)^{a+1/2}},
\end{aligned}
\label{eq:supp_branch_weights}
\end{equation}
and $\Omega_{ki}(D)=Z_{ki}^{\mathrm n}(D)/
(Z_{ki}^{\mathrm n}(D)+Z_{ki}^{\mathrm o}(D))$. Thus,
\begin{equation}
\begin{aligned}
\frac{Z_{ki}^{\mathrm n}(D)}{Z_{ki}^{\mathrm o}(D)}
&=C_i(b_i+D/2)^{a+1/2}e^{-u_iD/2},\\
C_i&=\frac{\theta e^{-v_i/2}\Gamma(a)}
{(1-\theta)b_i^a\Gamma(a+1/2)}>0.
\end{aligned}
\label{eq:supp_branch_ratio}
\end{equation}
Since $0\leq\Omega_{ki}(D)\leq
Z_{ki}^{\mathrm n}(D)/Z_{ki}^{\mathrm o}(D)$,
\begin{equation}
0\leq D\Omega_{ki}(D)
\leq C_iD(b_i+D/2)^{a+1/2}e^{-u_iD/2}
\xrightarrow[D\to\infty]{}0.
\end{equation}
The exponential decay dominates polynomial growth,
establishing \eqref{eq:supp_vanishing}. Since also
$\Omega_{ki}(D)\to0$, the shape increment
$\Omega_{ki}(D)/2$ and scale increment
$D\Omega_{ki}(D)/2$ both vanish.

This result also characterizes the direct influence
of an observation on the posterior mean variance.
Let $-$ and $+$ denote quantities before and after
including the selected observation in a single
variance update, respectively.
Consider a reference factor
$\operatorname{IG}(\alpha_i^-,\beta_i^-)$, with
$\alpha_i^->1$ and
$\hat s_i^-=\beta_i^-/(\alpha_i^--1)$.
Using \eqref{eq:supp_variance_update}, the corresponding
change in posterior mean variance is
\begin{equation}
\hat s_i^+-\hat s_i^-=
\frac{\Omega_{ki}(D)(D-\hat s_i^-)}
{2(\alpha_i^--1)+\Omega_{ki}(D)}
\xrightarrow[D\to\infty]{}0.
\label{eq:supp_variance_change}
\end{equation}
Hence, increasingly extreme observations have
vanishing direct influence on the posterior mean
nominal variance under fixed variational quantities.

\subsection{Covariance-Scaling Comparison}

To examine the role of decoupling in the above result,
consider extending the ASOR covariance-scaling likelihood
$\mathcal N(y_{ki};h_i(\mathbf x_k),s_i/I_{ki})$
with inverse-Gamma nominal-variance learning.
This is a hypothetical extension, since ASOR assumes
known nominal covariance. Let $I_{ki}=1$ on the nominal
branch and follow a Gamma prior with shape $a$ and
fixed rate $b_{\mathrm A,i}$ on the outlier branch.
Writing $\Omega_{\mathrm A}(D)$ for the nominal probability,
the expected precision multiplier is
\begin{equation}
\begin{aligned}
w_{\mathrm A}(D)
&=\Omega_{\mathrm A}(D)
+(1-\Omega_{\mathrm A}(D))
\frac{a+1/2}{b_{\mathrm A,i}+u_iD/2},\\
Dw_{\mathrm A}(D)&\xrightarrow[D\to\infty]{}
\frac{2a+1}{u_i}.
\end{aligned}
\label{eq:supp_asor_weight}
\end{equation}
The limit follows from the exponential decay of
$\Omega_{\mathrm A}(D)$ up to a polynomial factor,
which implies $D\Omega_{\mathrm A}(D)\to0$.

Unlike D-ASOR, both likelihood branches depend on
$s_i$ in the coupled model. Consequently, the reference
inverse-Gamma parameters change to $\alpha_i^-+1/2$
and $\beta_i^-+Dw_{\mathrm A}(D)/2$, yielding
\begin{equation}
\begin{aligned}
\hat s_{\mathrm A,i}^{+}-\hat s_i^-
&=\frac{Dw_{\mathrm A}(D)-\hat s_i^-}
{2\alpha_i^--1}\xrightarrow[D\to\infty]{}
\frac{(2a+1)/u_i-\hat s_i^-}{2\alpha_i^--1}.
\end{aligned}
\label{eq:supp_asor_variance}
\end{equation}
In contrast to \eqref{eq:supp_variance_change}, this
limit is generally nonzero. Thus, downweighting an
outlier in a coupled model does not necessarily
eliminate its direct influence on nominal-variance
learning. These results hold under fixed variational
quantities and do not establish global robustness
or convergence.

\section{Numerical Validation}
\label{sec:supp_numerics}

We isolate the effect of one reading on its channel's nominal
variance. In 30 paired Monte Carlo trials with Gaussian
contamination at $(\rho,p_{\rm out})=(10,0.4)$, both methods fit
the full 500-step record with the range reading from channel~1
at $k=250$ withheld. Let $\hat s_{M,1}^{-}$ be the fitted
variance for channel~1, where $M=\mathrm D$ denotes D-ASOR
and $M=\mathrm A$ denotes baseline ASOR. Let $r=\sqrt{D/\hat s_{\mathrm D,1}^{-}}\geq1$
denote the normalized expected residual magnitude.
For each $r$, both methods use
$D=r^2\hat s_{\mathrm D,1}^{-}$ and apply one variance
update with their fitted variational quantities fixed,
yielding $\hat s_{M,1}^{+}(r)$.
\begin{figure}[!ht]
\centering
\includegraphics[width=.96\linewidth,height=1.55in,
keepaspectratio]{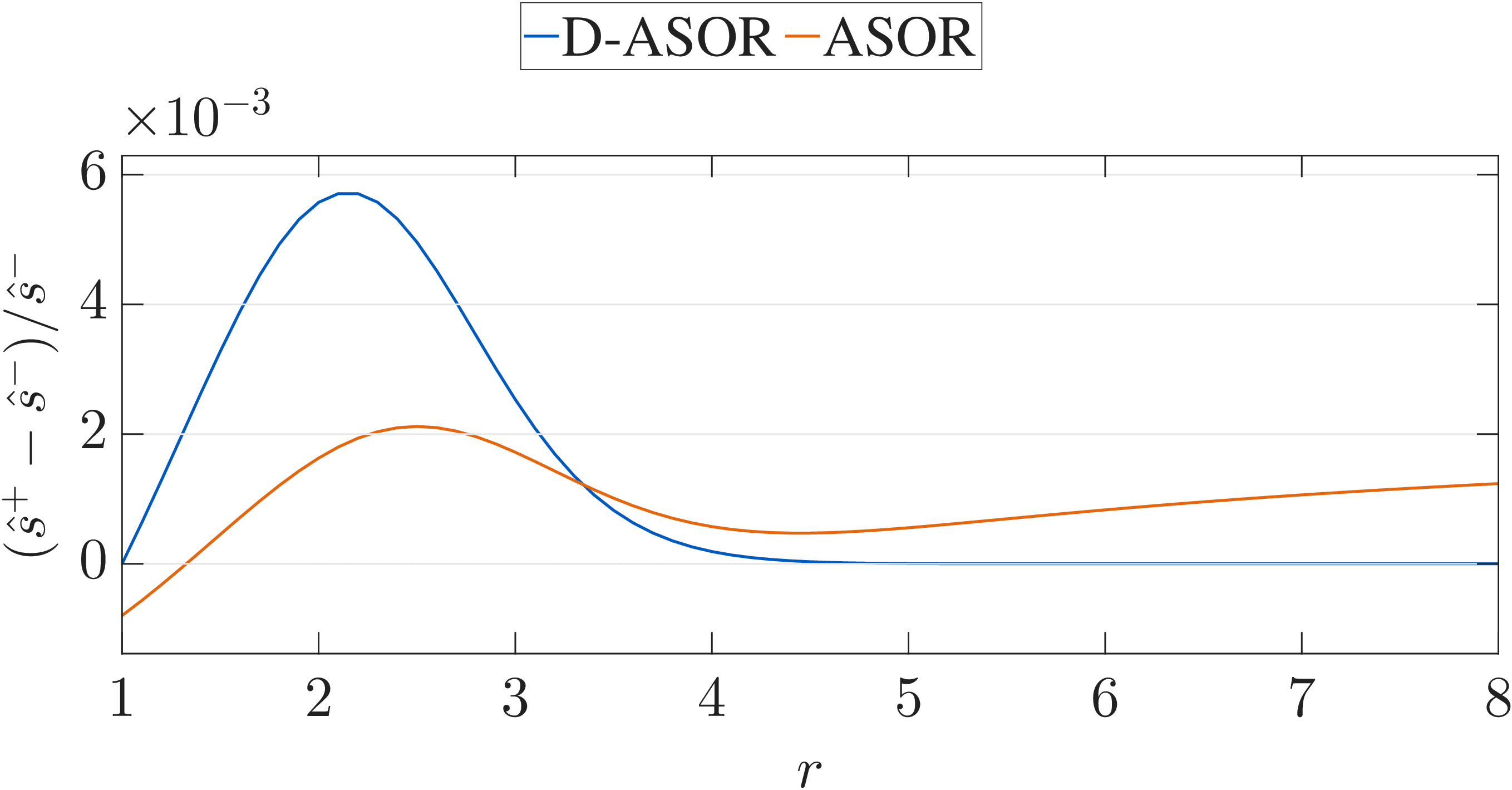}
\caption{Mean relative variance change
$(\hat s_{M,1}^{+}(r)-\hat s_{M,1}^{-})/\hat s_{M,1}^{-}$
over feasible trial pairs for each method $M$.}
\label{fig:supp_influence}
\end{figure}

At large $r$, Fig.~\ref{fig:supp_influence} shows negligible
variance change for D-ASOR and a persistent positive change
for coupled ASOR, consistent with \eqref{eq:supp_variance_change}
and \eqref{eq:supp_asor_variance}, respectively.


\clearpage
\setlength{\abovedisplayskip}{2pt plus 1pt minus 1pt}
\setlength{\belowdisplayskip}{2pt plus 1pt minus 1pt}
\setlength{\abovedisplayshortskip}{1pt plus 1pt}
\setlength{\belowdisplayshortskip}{2pt plus 1pt minus 1pt}
\setlength{\jot}{1pt}

\twocolumn[{\begin{center}
{\bfseries Information for Reproducibility}\\[-1pt]
{\small Outlier-Robust Variational Bayesian Smoothing with Decoupled Hierarchical Likelihoods}
\end{center}}]

\noindent\textit{Tracking and sensors:}\enspace
Experiments were conducted in MATLAB R2026a on a desktop
equipped with an AMD Ryzen 9 9950X processor (16 cores,
32 threads, 4.3~GHz base clock) and 64~GB of RAM.
Thirty paired Monte Carlo (MC) trials were used per
configuration, with identical data for all methods.
The coordinated-turn state is
$\mathbf x_k=[p_x,v_x,p_y,v_y,\omega]^\top$, with $d=1$~s.
With $\mathbf p=[p_x,p_y]^\top$ and
$\mathbf v=[v_x,v_y]^\top$, its noise-free dynamics are
\begin{equation}
\begin{aligned}
\mathbf p_k&=\mathbf p_{k-1}+\mathbf G(\omega_{k-1})\mathbf v_{k-1},\\
\mathbf v_k&=\mathbf R(\omega_{k-1})\mathbf v_{k-1},\quad
\omega_k=\omega_{k-1},\\
\mathbf R(\omega)&=\begin{bmatrix}\cos(\omega d)&-\sin(\omega d)\\
\sin(\omega d)&\cos(\omega d)\end{bmatrix},\\
\mathbf G(\omega)&=\frac{1}{\omega}\begin{bmatrix}
\sin(\omega d)&\cos(\omega d)-1\\1-\cos(\omega d)&\sin(\omega d)
\end{bmatrix}.
\end{aligned}
\label{eq:repro_transition}
\end{equation}
At $\omega=0$, $\mathbf R=\mathbf I_2$, $\mathbf G=d\mathbf I_2$.
The known process-noise covariance is
\begin{equation}
\begin{gathered}
\mathbf Q=\operatorname{blkdiag}(0.1F_d,0.1F_d,1.75\times10^{-4}),\\[-2pt]
F_d=\begin{bmatrix}d^3/3&d^2/2\\d^2/2&d\end{bmatrix}.
\end{gathered}
\label{eq:repro_process}
\end{equation}
Set $\mathbf x_1=\mathbf m_1=[0,10,0,-5,-3\pi/180]^\top$,
$\mathbf P_1=\mathbf Q$, $K=500$ and $m=10$ (five ranges and
five bearings). Sensor locations in metres are
\begin{equation}
\begin{aligned}
\mathbf r_j&=350(j-1,\operatorname{mod}(j+1,2)),\\
\mathbf b_j&=350(j-1,\operatorname{mod}(j,2)),\quad j=1,\ldots,5.
\end{aligned}
\label{eq:repro_sensors}
\end{equation}
The observation map uses Euclidean ranges and $\operatorname{atan2}$
bearings. Nominal standard deviations are $\sqrt{10}(1,2,1,2,1)$~m
and $0.2^\circ(1,2,1,2,1)$, respectively, initialized by
non-oracle methods at $5$~m and $1^\circ$. Bearings are in radians,
with angular differences wrapped to $[-\pi,\pi)$.

\noindent\textit{Contamination:}\enspace
The nominal noise, outlier amplitudes, and indicators are
mutually independent across channels and time, with
\begin{equation}
\begin{gathered}
y_{ki}=h_i(\mathbf x_k)+\varepsilon_{ki}+z_{ki}o_{ki},\\
\varepsilon_{ki}\sim\mathcal N(0,\sigma_i^2),\quad
z_{ki}\sim\operatorname{Bern}(p_{\rm out}),\\
o_{ki}\sim\begin{cases}
\mathcal N(0,\rho^2\sigma_i^2),&\text{Gaussian},\\
\mathcal U(0,\sqrt{12}\rho\sigma_i),&\text{uniform}.
\end{cases}
\end{gathered}
\label{eq:repro_contamination}
\end{equation}
We test $\rho\in\{5,10,15,20\}$ and
$p_{\rm out}\in\{0.2,0.4,0.6\}$.
Both outlier distributions have standard deviation
$\rho\sigma_i$; the uniform distribution has mean
$\sqrt3\rho\sigma_i$.

\noindent\textit{Methods:}\enspace
VB-HAUKF and projected-diagonal VBAKF-GGM use iterative
RTS smoothing with smoothed residual statistics. Alongside
D-ASOR, fixed-covariance and oracle UKF--RTS are compared.
For the nominal variances $s_i=\sigma_i^2$, matched priors
$s_i\sim\operatorname{IG}(\alpha_{0i},\beta_{0i})$
use $\alpha_{0i}=2.1$ and
$\beta_{0i}=\sigma_{0i}^2[\Gamma(\alpha_{0i})/\Gamma(\alpha_{0i}-1/2)]^2$,
so $\mathbb E_0[\sigma_i]=\sigma_{0i}$. D-ASOR uses
$\theta=a=0.5$, $b_i^{(0)}=50$, $A=2$, $B_{0i}=1/50$.
The unscented transform parameters are
$(\alpha_{\rm UT},\beta_{\rm UT},\kappa)=(1,2,0)$.
VB-HAUKF uses Huber threshold $1.345$ and forgetting factor
$1-e^{-3}$; VBAKF-GGM uses $1-e^{-4}$, four Gamma components
(shape $10$, rates $10,100,1000,10000$), and unit Dirichlet parameters.

\noindent\textit{Oracle:}\enspace
Knowing the true nominal variances, outlier indicators and
contamination moments, the oracle subtracts $z_{ki}\mathbb E[o_{ki}]$
and uses variance $\sigma_i^2(1+z_{ki}\rho^2)$.
It serves as a moment-informed reference rather than
a theoretical lower bound.

\newpage

\noindent\textit{Stopping and accuracy:}\enspace
Convergence is assessed by dividing the $\ell_2$ norm of the
change in stacked smoothed state means by the $\ell_2$ norm
of the previous stacked estimate.
Iterations terminate when the ratio falls below $10^{-4}$
or after 100 iterations. Finite outputs at the iteration
cap are retained. D-ASOR repeats the forward and backward
passes with final precisions and estimates
$\hat\sigma_i=\sqrt{\beta_i/(\alpha_i-1)}$.
For $M=30$, $\mathcal E=\{151,\ldots,500\}$,
$N=350$, and $n=5$, define
\begin{equation}
\begin{aligned}
E_x&=\sqrt{\frac{1}{MNn}\sum_{j=1}^{M}\sum_{k\in\mathcal E}\sum_{\ell=1}^{n}
\left(\frac{m_{k,\ell}^{s,(j)}-x_{k,\ell}^{(j)}}{c_\ell}\right)^2},\\
c_\ell&=\sqrt{\frac{1}{MN}\sum_{j=1}^{M}\sum_{k\in\mathcal E}
(x_{k,\ell}^{(j)})^2}.
\end{aligned}
\label{eq:repro_metrics}
\end{equation}
All methods share the ground-truth RMS scales $c_\ell$
within each configuration. The dimensionless state error
$E_x$ excludes the first 30\% of each trajectory.

\noindent\textit{Measurement reconstruction:}\enspace
In trial $j$, reconstruct each measurement as
$\hat y_{ki}^{(j)}=h_i(\mathbf m_k^{s,(j)})$ and compute
\begin{equation}
E_y^{(j)}=\sqrt{\frac{1}{m(K-1)}\sum_{k=2}^{K}\sum_{i=1}^{m}
\left(\frac{h_i(\mathbf x_k^{(j)})-\hat y_{ki}^{(j)}}{\sigma_i}\right)^2}.
\label{eq:repro_reconstruction}
\end{equation}
Angular differences are wrapped to $[-\pi,\pi)$.
Fig.~1(a) reports the metric over 30 trials under
uniform contamination at $\rho=10$.

\noindent\textit{Computational cost:}\enspace
Fig.~1(b) reports runtimes over 30 paired trials for
$m=50,100,\ldots,250$ under Gaussian contamination
with $\rho=10$ and $p_{\rm out}=0.2$,
using $m/2$ sensors of each type.
For runtime experiments, the sensor layout in
\eqref{eq:repro_sensors} is extended to
$j=1,\ldots,m/2$ for each measurement type.
For centered state and measurement sigma-point deviation
matrices $\mathbf X_k$ and $\mathbf Y_k$, weighted by the
square roots of covariance weights, with $L=2n+1$
columns, Woodbury's identity gives
\begin{equation}
\begin{aligned}
\mathbf K_k&=\mathbf U_k\mathbf S_k^{-1}=\mathbf X_k(\mathbf I_L+
\mathbf Y_k^\top\mathbf T_k\mathbf Y_k)^{-1}\mathbf Y_k^\top\mathbf T_k.
\end{aligned}
\label{eq:repro_woodbury}
\end{equation}
Here $\mathbf U_k=\mathbf X_k\mathbf Y_k^\top$ and
$\mathbf S_k=\mathbf T_k^{-1}+\mathbf Y_k\mathbf Y_k^\top$.
Diagonal $\mathbf T_k$ makes the reduced solve, including
assembly, $\mathcal O(mL^2+L^3)$ per step, hence
$\mathcal O(m)$ per iteration for fixed $n,K$.
Runtimes were measured using single-threaded MATLAB execution.

\noindent\textit{Prior sensitivity:}\enspace
In Fig.~1(c), we use Gaussian contamination with
$(\rho,p_{\rm out})=(10,0.4)$.
For range channels $1,3,5$ with $\sigma_i^\star=\sqrt{10}$~m,
we vary $\mathbb E_0[\sigma_i]\in\{1,2,4,\ldots,40\}$~m and
$\operatorname{Var}_0(\sigma_i)\in\{1,5,10,\ldots,100\}$~m$^2$.
The inverse-Gamma parameters ($\alpha_{0i}>1$) are
determined by matching the moments
\begin{equation}
\begin{aligned}
\mathbb E_0[\sigma_i]&=\sqrt{\beta_{0i}}
\frac{\Gamma(\alpha_{0i}-1/2)}{\Gamma(\alpha_{0i})},\\
\operatorname{Var}_0(\sigma_i)&=\frac{\beta_{0i}}{\alpha_{0i}-1}
-\big(\mathbb E_0[\sigma_i]\big)^2.
\end{aligned}
\label{eq:repro_prior_moments}
\end{equation}
Set $\sigma_{0i}=\mathbb E_0[\sigma_i]$ for these channels and keep
the other seven priors fixed. The same 30 datasets are reused
at every grid point. The surface reports the mean trial-level
nominal-noise NRMSE
\begin{equation}
\mathrm{NRMSE}_\sigma=\frac{1}{M}\sum_{j=1}^{M}
\sqrt{\frac{1}{3}\sum_{i\in\{1,3,5\}}
\left(\frac{\hat\sigma_i^{(j)}-\sigma_i}{\sigma_i}\right)^2}.
\label{eq:repro_prior_nrmse}
\end{equation}

\end{document}